\documentclass{article}

\usepackage[final,nonatbib]{nips_2016}

\usepackage[utf8]{inputenc} % allow utf-8 input
\usepackage[T1]{fontenc}    % use 8-bit T1 fonts
\usepackage{hyperref}       % hyperlinks
\usepackage{url}            % simple URL typesetting
\usepackage{booktabs}       % professional-quality tables
\usepackage{amsfonts}       % blackboard math symbols
\usepackage{nicefrac}       % compact symbols for 1/2, etc.
\usepackage{microtype}      % microtypography

\usepackage{amssymb,amsmath,epsfig}
\usepackage{graphicx}
\usepackage{times} 
\usepackage{color}
\usepackage{wrapfig}
\usepackage{epsf}
\usepackage{colortbl}
\usepackage{verbatim}
\usepackage{subfigure}
\usepackage{multirow,rotating}
\usepackage{fancyheadings} 
\usepackage{xcolor} 
\usepackage{color, colortbl}
\usepackage{amstext}
\usepackage{paralist}
\usepackage{enumitem}
\usepackage{pdfpages}

\usepackage{algorithm}
\usepackage{algpseudocode}
\usepackage{amssymb}
\usepackage{amsmath}
\usepackage{lscape}

\title{Deep Reinforcement Learning for \\Multi-Domain Dialogue Systems\thanks{Funding from Samsung Electronics Ltd. and the University of Lincoln is gratefully acknowledged. We thank Raymond Kirk for helping with App development (to integrate the agents in this paper) and system testing. We also thank Heesik Jeon and the AI team at Samsung for their executive efforts in this project.}}

\author{
  Heriberto Cuay\'ahuitl$^1$, Seunghak Yu$^2$, Ashley Williamson$^1$, Jacob Carse$^1$  \\
  $^1$University of Lincoln, School of Computer Science, United Kingdom\\
  $^2$Artificial Intelligence Team, Samsung Electronics Co. Ltd., Seoul, South Korea \\
  \texttt{HCuayahuitl@lincoln.ac.uk} \\
}

\begin{document}

\maketitle

\begin{abstract}
Standard deep reinforcement learning methods such as Deep Q-Networks (DQN) for multiple tasks (domains) face scalability problems. We propose a method for multi-domain dialogue policy learning---termed NDQN, and apply it to an information-seeking spoken dialogue system in the domains of restaurants and hotels. Experimental results comparing  DQN (baseline) versus NDQN (proposed) using simulations report that our proposed method exhibits better scalability and is promising for optimising the behaviour of multi-domain dialogue systems.
\end{abstract}

\section{Introduction}
Dialogue systems based on the Reinforcement Learning (RL) paradigm offer the possibility to treat dialogue design as an optimisation problem, and are attractive because they can improve their performance over time with experience. But the application of RL is not trivial due to the complexity of the problem such as large state-action spaces exhibited in human-machine conversations. This is especially true in multi-domain systems, where the number of state variables (features) and dialogue actions increases rapidly as more domains are taken into account. On the one hand, unique situations in the interaction can be described by a large number of variables (e.g.\ words raised in the conversation by the system and user) so that enumerating them would result in very large state spaces. On the other hand, the action space can also be large due to the wide range of unique dialogue actions (e.g.\ requests, apologies, confirmations in multiple contexts).

While one can aim to optimise the interaction via compression of the search space, it is usually unclear what features to incorporate in the state representation. This is a strong motivation for applying Deep Reinforcement Learning (DRL) to dialogue management so that the agent can simultaneously learn its feature representation and policy \cite{mnih-atari-2013}. This paper makes use of raw noisy text as features in an attempt to avoid engineered features to represent the dialogue state. By using this representation, dialogue agents bypass spoken language understanding in order to learn dialogue policies directly from raw (noisy) text to actions \cite{Cuayahuitl16}.

We address dialogue optimisation using the divide-and-conquer approach, in which dialogue states can be described at different levels of granularity, and an action can execute behaviour using either a single dialogue action (taking one dialogue turn) or a composite one (equivalent to a subdialogue taking multiple dialogue turns). This approach offers at least two benefits: (a) modularity helps to optimise subdialogues that may be easier to optimise than the whole dialogue; and (b) subdialogues may include only relevant dialogue knowledge in the states and relevant actions, thus reducing significantly the size of possible solutions: consequently they can be found faster. These properties are crucial for training the behaviour of multi-domain spoken dialogue systems in which there may be a large set of state variables or a large number of actions.

Below we describe a data-driven method to the approach described, which we have applied to an information-seeking dialogue system in the domains of restaurants and hotels. Experimental results show that the proposed method can train policies faster and more effectively than a standard algorithm in the literature, showing promise for training multi-domain dialogue systems.

\section{Literature Review}
%Multi-domain dialogue systems: Komatani2009, CuayahuitlRLS10, Lison11, GasicMSVWY15, 
Recently, multi-domain spoken conversational agents have received an increasing amount of attention. This may be due to the fact that speech technologies such as Automated Speech Recognition (ASR) and Text-To-Speech (TTS) have reached a high degree of maturity. But the question of {\it How to design conversational systems for human-machine interaction in multiple domains (or tasks)?} is still an open and interesting problem in artificial intelligence. The dialogue system proposed by \cite{Komatani2009} used a distributed architecture of domain experts modulated by a domain selector. The latter used a decision tree with classification errors over 20\% in 5 domains. This indicates that not only individual domains have to exhibit robust interactions against errors, but also that errors increase by incorporating more domains.\cite{JeonOHK16} used rule-based classifiers for predicting user intentions, which are executed using a Hierarchical Task Network (HTN) incorporating expert knowledge. Trainable multi-domain dialogue systems using traditional reinforcement learning include \cite{CuayahuitlRLS10, Lison11, WangCWTWW14, GasicMSVWY15}. These systems use a modest amount of features, and  in contrast to neural-based systems, they require manual feature engineering.

%Neural dialogue systems: CuayahuitlKL15, SerbanSBCP15, GeXu2015, VandykeSGMWY15, MrksicSTGSVWY15, SerbanSBCP15, VinyalsL15, WestonCB14
Recent work on deep learning applied to task-oriented conversational agents include the following. \cite{GeXu2015} uses a Recurrent Neural Network (RNN) for dialogue act prediction in a POMDP-based dialogue system, which focuses on mapping system and user sentences to dialogue acts. \cite{CuayahuitlKL15} applies Deep Reinforcement Learning with a fully-connected neural network for trading negotiations in board games, which focuses on mapping game situations to dialogue actions. \cite{VandykeSGMWY15} trains RNN-based classifiers for predicting dialogue success in multi-domain dialogue systems, which can be applied to unseen domains. \cite{MrksicSTGSVWY15} also trains RNN-based classifiers but for belief tracking in order to improve the robustness of recognised user responses across dialogue turns. Other neural-based conversational agents have been applied to text prediction using the sequence-to-sequence approach \cite{SerbanSBCP15,VinyalsL15}, and to reasoning with inference for text-based question answering \cite{WestonCB14}. 

From these works, we can observe that supervised learning is the dominating form of training in neural-based conversational agents. To our knowledge, we report the first multi-domain dialogue system using deep reinforcement learning. This form of learning is interesting because it can perform feature learning and policy learning simultaneously, and its effective application in real-world dialogue scenarios remains to be demonstrated. 

\section{Method}
\label{sec:method}
Our proposed method to scale up Deep Reinforcement Learning (DRL) for multi-domain dialogue systems has two stages. First, multi-policy learning via a network of DRL agents; and second, more compact state representations by compressing raw inputs. Although these two stages can be applied independently, their combination aims for further scalability than any one of them individually.

\subsection{Network of Deep Q-Networks}
\label{sec:ndqn}
\begin{figure}[h!]
\begin{center}\centerline{\includegraphics[scale=0.53]{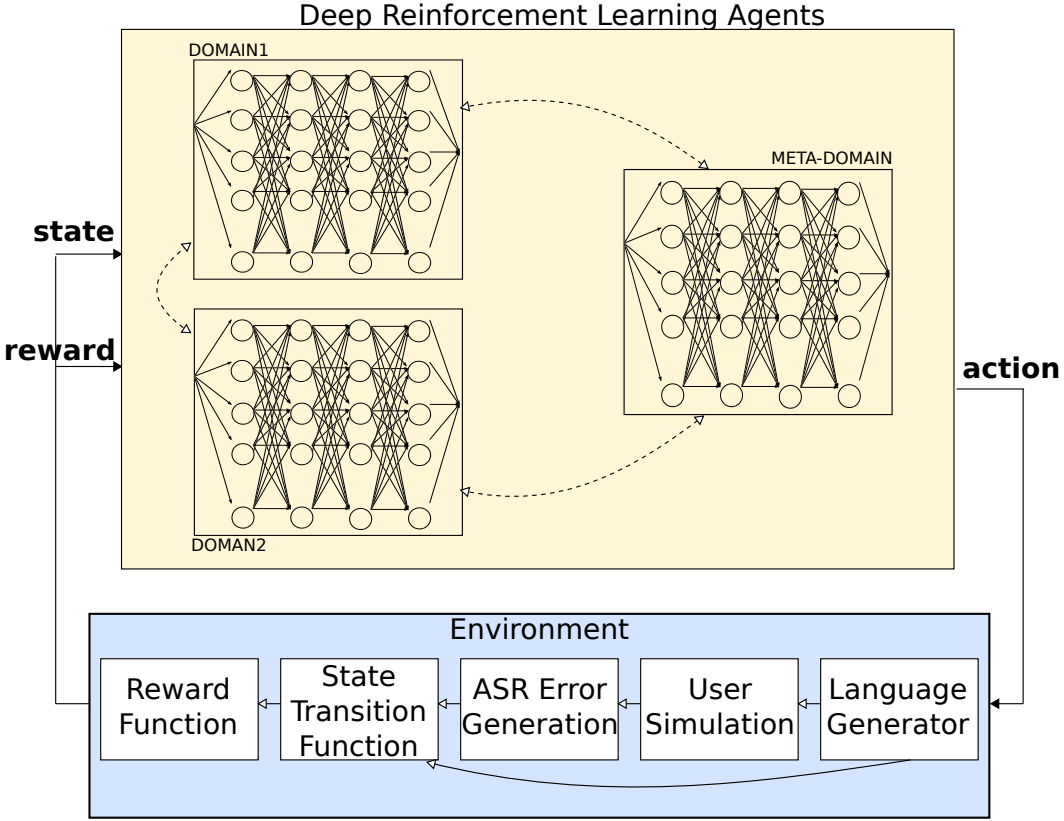}}
\end{center}
\caption{Multi-domain DRL agents with flexible interaction. 
The dashed arrows connecting domains denote flexible transitions between domains in order to avoid a rigid structure in the interaction. Although all policies are considered for decision-making, only one domain can be executed at a time implying that a previous domain can continue its execution in order to resume the interaction.}
\label{multidsArchitecture}
\end{figure}

We propose to optimise multi-domain dialogue systems using a network of Deep Reinforcement Learners, for example a network of Deep Q-networks (DQN) --- see \cite{mnih-atari-2013,mnih-dqn-2015} for an introduction to the standard DQN method. In our method, instead of training a single DQN,we train a set of DQNs (also referred to as NDQNs), where every DQN represents a specialised skill to converse in a particular subdialogue --- see Figure~\ref{multidsArchitecture}. In addition, the network of agents enable DQNs to be executed without a fixed structure in order to support flexible and unstructured dialogues. In contrast to hierarchical DQNs \cite{KulkarniNST16} that follow a strict sequence of agents, in our method an NDQN allows transitions between all DQN agents except for self-transitions and loops (the latter using a stack-based approach as in \cite{Cuayahuitl2014tiis}). Furthermore, while user responses can motivate transitions to another domain in the network, completing a subdialogue within a domain motivates a transition to the previous domain to resume the interaction. Algorithm~\ref{ndqn} describes the procedure to train and execute NDQNs.

An optimal policy in an NDQN performs action selection according to
\begin{equation}
\pi^{*}_{\theta^{(d)}}(s) = \arg \max_{a \in A^{(d)}} Q^{*(d)}(s,a;\theta^{(d)}), %\mbox{ where } d \in D.
\label{eq:policyRLNs}
\end{equation}
where domain or skill $d \in D$ is selected according to 
\begin{equation}
d = \arg \max_{d' \in D} F(d'|d,{\bf e}), 
\label{eq:domainTrans}
\end{equation}
and evidence {\bf e} takes into account all features that describe the environment space of domain $d$. While this transition function (Eq.~\ref{eq:domainTrans}) is used for high-level transitions in the interaction, Equation~\ref{eq:policyRLNs} is used for low-level transitions within a node (skill) in the network and subject to reinforcement learning. NDQN assumes that the domain transition function $F$ can be deterministic or probabilistic (the latter due to uncertainty in the interaction), and it is a prior requirement for NDQN-Learning.
%, and that DQN agents 
%make use of a universal or  reward function or independent ones. 

\begin{algorithm} [t]
\label{ourAlgorithm}
\caption{Network of Deep Q-Learners (NDQN)}\label{ndqn} 
\begin{algorithmic}[1]
\State Initialise set of Deep Q-Networks with replay memories $D^{(d)}$, action-value functions $Q^{(d)}$ with random weights $\theta^{(d)}$, and target action-value functions $\hat{Q}^{(d)}$ with weights $\hat{\theta}^{(d)}=\theta^{(d)}$
%\For{episode=1 to M}
\Repeat
   \State $d \leftarrow$ initial domain, predefined or defined by $\arg\max_{d \in D} F_o(d)$
   \State $s \leftarrow$ initial environment state in $S^{(d)}$
   \Repeat 
   \Repeat
      \State Choose action $a \in A^{(d)}$ in $s$ derived from $Q^{(d)}$ (e.g. $\epsilon$-greedy, Thompson)
      \State Execute action $a$ and observe reward $r$ and next state $s'$
      \State Append transition ($s,a,r,s'$) to $D^{(d)}$
      \State $B^{(d)} \leftarrow$ sample random minibatch of experiences from $D^{(d)}$
      \State $d' \leftarrow \mbox{select next domain according to }\arg \max_{d' \in D} F(d'|s',{\bf e})$
      \State $y_j= 
\begin{cases}
    r_j& \text{if } \mbox{final step of episode}\\
    r_j + \gamma \max_{a \in A^{(d)}} \hat{Q}^{(d)}(s',a';\hat{\theta}^{(d)}),              & \text{otherwise}
\end{cases}$
%\text{if }n=n'\\
%    r_j + \gamma \max_{a \in A^{(n')}} \hat{Q}^{(n')}(s',a';\hat{\theta}^{(n')}),              & 
      \State Gradient descent step on $\left( y_j-Q^{(d)}(s',a';\theta^{(d)}) \right)^2$ using $B^{(d)}$
      \State Reset $\hat{Q}^{(d)}=Q^{(d)}$ every $C$ steps
      \State $s \leftarrow$ $s'$
%      \If {$n$ = $n'$} % or $s$ is a terminal state}
%         \State {\bf break}
%      \EndIf
   \Until {$s$ is a terminal state or $d \neq d'$}
   \State $d \leftarrow$ $d'$
\Until {$s$ is a goal state}
\Until convergence
%\EndFor
\end{algorithmic}
\end{algorithm}

\subsection{NDQNs with Compressed Raw Inputs}
Previous work on dialogue policy learning using DRL map raw (noisy) text to actions \cite{Cuayahuitl16,ZhaoE16}. This is not only computationally intensive, but it becomes infeasible for dialogue systems with large vocabularies. To tackle this problem we propose to use delexicalised sentences (as proposed in \cite{HendersonTY14}) and synonymised sentences. This has the advantage that dialogue policies can be trained from more compact state representations than those using only raw inputs, and have coverage for a larger vocabulary than trained for.

\subsubsection{Delexicalisation}
%Regarding {\bf delexicalisation}, c
Consider a dialogue system for restaurant search receiving the  following user request---with corresponding delexicalised sentence underneath. 
\begin{figure}[h!]
\begin{center}\centerline{\includegraphics[scale=0.75]{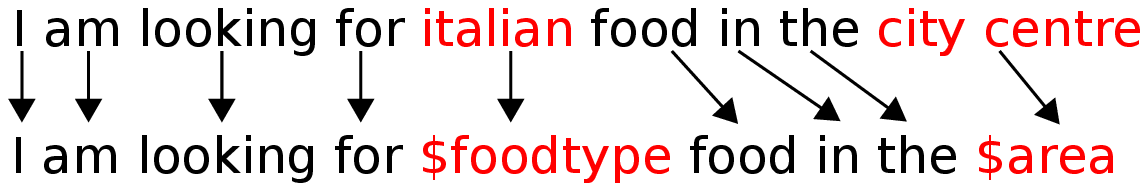}}
\end{center}
%\caption{Example user request (top) with replaced slot values (bottom)}
\end{figure}
%{\it I am looking for Italian food in the city centre}. A delexicalised version of this sentence corresponds to  
%\begin{verbatim}
%I am looking for $foodtype food in the $area.
%\end{verbatim}
\vskip-10pt
The latter representation combining words and slot IDs (denoted with the symbol `\$') has several practical advantages. For example, policies can be learnt faster, they contribute to further scalability of systems with large vocabularies, and policies do not have to be retrained if the slot values change over time. In this work we use heuristics to replace slot values by slot IDs, and a trainable component for automatic slot labelling is considered beyond the scope of this paper. 
%as in \cite{Cuayahuitl2014slt,BengioEtAl2015}
%The cost of this advantages requires training a classifier (or classifiers) to predict the slot values in the inserted slot(s). 

\subsubsection{Synonymization}
%Regarding {\bf synonymization}, c
%{\it I am hoping for a nice pasta in the city centre}---
Consider the same system above receiving the following user request given the unknown words `fancy' and `cuisine'---with corresponding synonyms underneath.
\begin{figure}[h!]
\begin{center}\centerline{\includegraphics[scale=0.75]{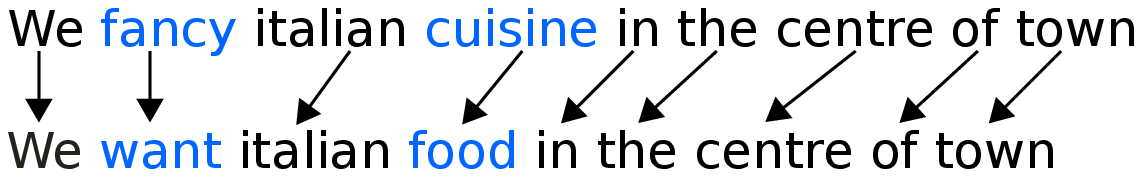}}
\end{center}
%\caption{Example user request (top) with replaced known words (bottom)}
\end{figure}
\vskip-10pt
We argue that word synonyms can be useful in such  situations because the unknown word `fancy' can trigger the known word feature `want'. Similarly, the unknown word `cuisine' can trigger the known word feature `food'. In this way, the vocabulary of our NDQNs incorporate a mapping from filler words and slot values to synonyms in order to cope with unseen wordings. Due to the complexity of automatic generation of meaningful synonyms, our synonyms have been manually specified but they can be generated automatically for example from word embeddings  \cite{MikolovSCCD13}.

\section{Multi-Domain Dialogue System}
\label{sec:multids}
The proposed computational framework for training multi-domain dialogue agents is a substantial extension from the publicly available software tools SimpleDS \cite{Cuayahuitl16} and ConvnetJS \cite{ConvNetJS}. 
%, and is privately available at {\url{https://github.com/LCAS/MultiDS}}. 
It can be executed in training or test mode using simulations or speech-based interactions (via an App). Our dialogue system runs under a client-server architecture, where the learning agents---one per domain---act as the {\it clients} and the dialogue system as the {\it server}. They communicate by exchanging messages, where the clients tell the server the action to execute, and the server tells the clients the state and reward observed. The elements for training multi-domain dialogue systems are as follows. 

\paragraph{State Spaces}
They include word-based features depending on the vocabulary of each learning agent. They include 177 unique words\footnote{The unique words in our system's vocabulary excludes words from information presentation due to the vast amount of information about hotels and restaurants. Nonetheless and during testing, our system retrieves live information from \url{http://www.bookatable.co.uk}) and \url{www.reservetravel.com}.} without synonyms, and 150 unique words with synonyms. For example, an agent in the domain of restaurants has relevant features for its domain and it is agnostic of features in other domains. While words derived from system responses are treated as binary variables (i.e. word present or absent), the words derived from noisy user responses can be seen as continuous variables by taking ASR confidence scores into account. Since a single variable per word is used, user features override system ones in case of overlaps. %Alternative state representations including part-of-speech tags, synonyms and word embeddings is work in progress.

\paragraph{Action Spaces}
They include dialogue acts for the targeted domains---currently 69 unique actions in total. Example dialogue act types, dialogue acts without parameters, are as follows: Salutation(), Request(), AskFor(), Apology(), ExpConfirm(), ImpConfirm(), Retrieve(), Provide(), among others. Rather than learning with whole action sets, our framework supports learning from constrained actions by applying learning updates only on the set of valid actions. These actions are derived from the most likely actions, $Pr(a|s)>0.0001$, from Naive Bayes classifiers (due to scalability purposes) trained from example dialogues. See example demonstration dialogue in Appendix~\ref{App:AppendixA}. In addition to the most probable data-like actions, the constrained actions are extended with the legitimate requests, apologies and confirmations. The fact that constrained actions are data-driven and driven by domain-independent heuristics, facilitates its usage across domains. 

\paragraph{State Transition Functions}
They are based on numerical vectors representing the last system and user responses. Taking a wider dialogue context is also possible but not explored in this paper. The system responses are straightforward, 0 if absent and 1 if present (hit-or-miss). The user responses correspond to the confidence level [0..1] of noisy user responses. While system responses are generated from stochastic templates, user responses are generated from semi-random user behaviour. These elements enable the creation of a vast amount of different conversations for agent training.

\paragraph{Domain Transition Function}
This function specifies the next domain or task in focus. It is currently defined deterministically, and it is also implemented as a SVM classifier trained from example interactions---see Appendix~\ref{App:AppendixA}. The design of this classifier follows that of a two-deep fully connected neural network with 80 nodes in each hidden layer, with tanh activation, and an SVM output layer, using Hinge Loss. While the input layer accepts domain-independent \textit{words-as-features} vectors representing the unique global vocabulary shared amongst all domains in a hit-or-miss approach, the output layer has 3 classes representing system domains (meta\footnote{We refer to meta domain as subdialogues containing domain-general system and user responses.}, restaurants and hotels). 15K dialogues of data were generated, partitioned as a 60-40 training-testing split, and trained for 180 epochs. Initial results of this classifier shows a 87.5\% classification accuracy on user-simulated data.
%150000 samples where randomly selected, with replacement, from the testing set to obtain classification accuracy.
%PLEASE BE BREAF HERE -- NO NEED TO EXPLAIN WHAT IS WELL KNOWN
%Attempted to touch on use of gVec as input to the domain transition function, as the SVM can't just use 'state-space' as this is domain-specific. It must use domain-independent feature vectors.
% CAN YOU CONFIRM IF THE SUPPORT-VECTOR NETWORKS REFERENCE IS REQUIRED. IT APPEARS TO FIT CLOSER THAN TYPICAL SVMS AS CONVNETJS IMPLEMNT IT AS A NN WITH WEIGHTS CHANGED ACCORDING TO HINGE LOSS.
% UNSURE ABOUT WHERE TO PLACE TRAIN/TESTING METHODOLOGY.

\paragraph{Reward Function}
It is defined as $R(s,a,s')=GR + DR - DL$, where $GR$ is goal-based reward treated as task success [0..1] (the proportion of positively confirmed slots and information retrieved and presented); $DR$ is a data-like probability of having observed action $a$ in state $s$; and $DL=t*w$ is a dialogue length measure used to encourage efficient interactions with $t$ time steps and weight $w$ (-0.1 in our case). The $DR$ scores are derived from Naive Bayes classifiers to allow statistical inference over actions given states ($Pr(a|s)$).

\paragraph{Model Architectures}
We use fully-connected multilayer neural nets, trained with stochastic gradient descent, where nodes in the input layers depend on the vocabulary of each agent. The use of convolutional neural nets is work in progress. They include 2 hidden layers with 80 nodes with Rectified Linear Units to normalise their weights \cite{NairH10}. Dropout\cite{GoodfellowWMCB13} and adaptive learning rates are also part of our work in progress. Other hyperparameters include experience replay size=10000, burning steps=1000, discount factor=0.7, minimum epsilon=0.001, batch size=32, and learning steps=30000.

\section{Experimental Results}
\label{sec:evaluation}
%This paper is motivated by the research question {\it ``How to train multi-domain dialogue systems with flexible interaction using deep reinforcement learning?''}. 
Here we compare a multi-domain dialogue system using a standard DRL method versus our proposed method described in Sections~\ref{sec:method} and~\ref{sec:multids}. While the former (DQN) uses a single policy for learning ({\it baseline}), the latter (NDQN) uses multiple  policies ({\it proposed}). Both multi-domain dialogue systems use the same data, resources and hyperparameters for training, the only difference between both systems is the learning method (DQN or NDQN) or state representation (with or without compression).

%\begin{figure}[h!]
%\begin{center}\centerline{\includegraphics[scale=0.4]{./multids-output-omni-oct7}}
%\end{center}
%\caption{Learning curves of the baseline system (with input compression)}
%\label{omni-single-policy}
%\end{figure}

\begin{figure}%[ht!]
     \begin{center}
%
        %\subfigure[Meta Domain: (left) without input compression, (right) with input compression]{%
            %\label{fig:first}
            \includegraphics[width=0.5\textwidth]{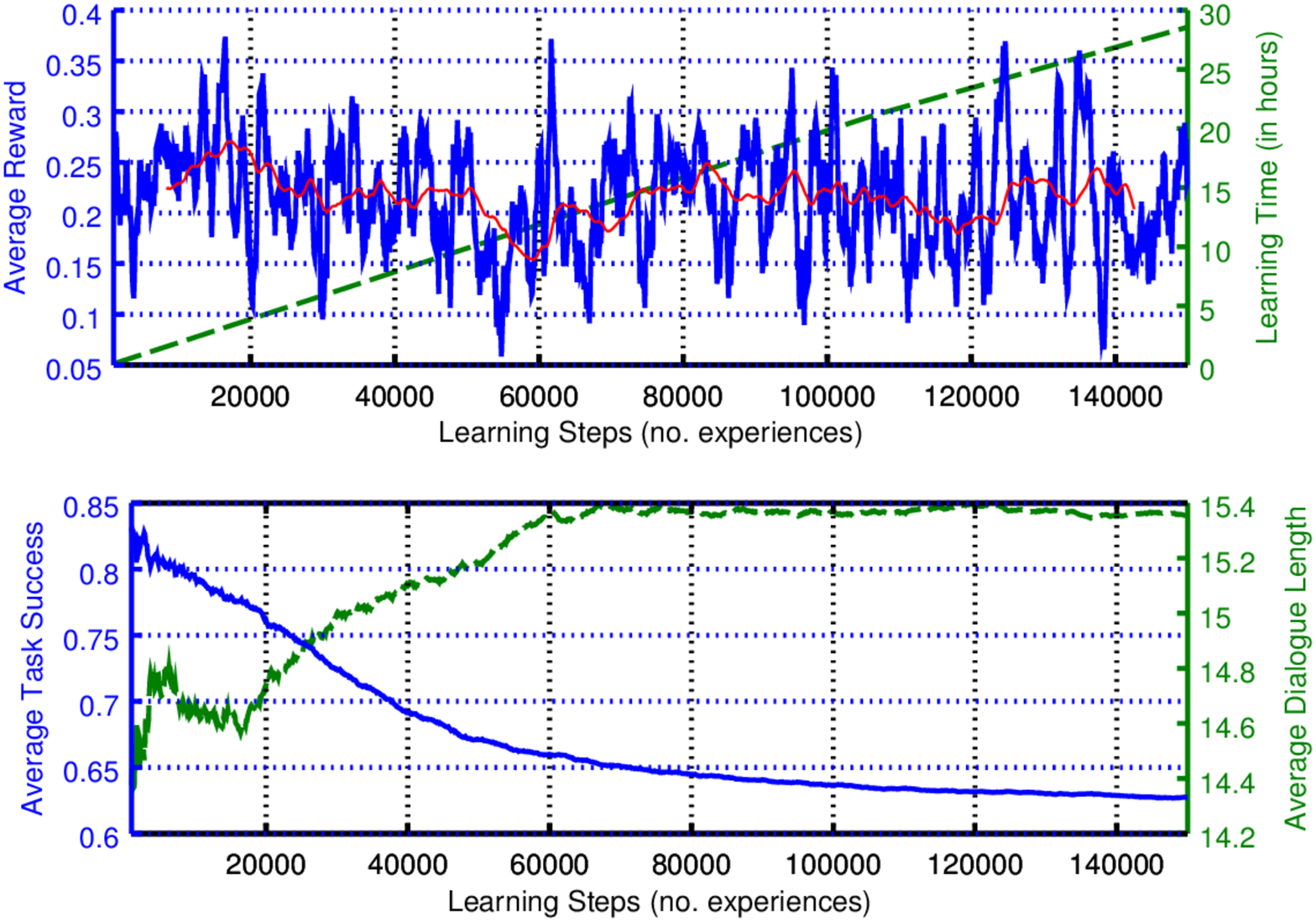}
		\hspace{-0.23cm}
            \includegraphics[width=0.5\textwidth]{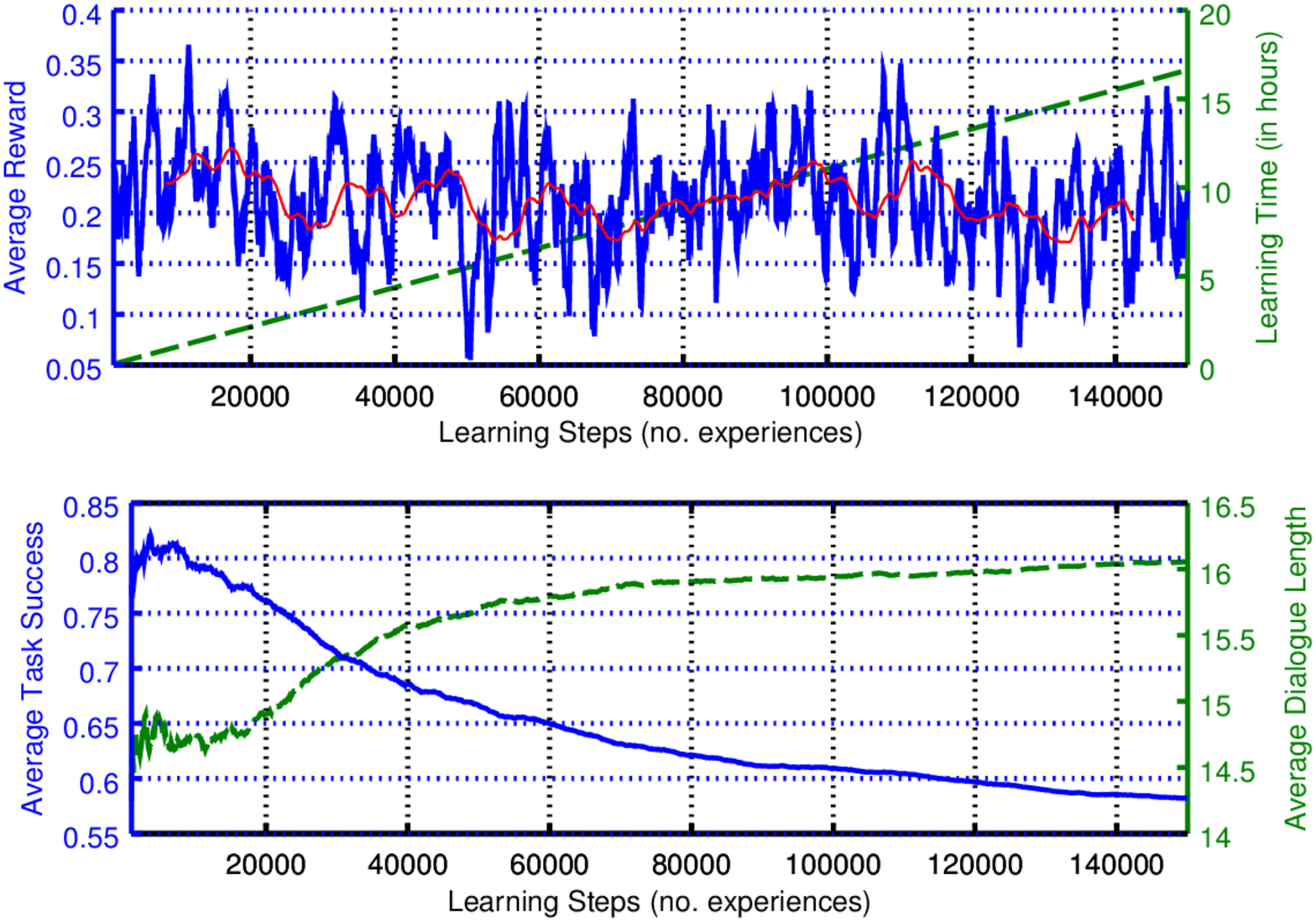}
        %}
%
    \end{center}
%        \vskip20pt
   \label{fig:singlePolicy}
   \caption{Learning curves of the baseline system: (left) without input compression, (right) with input compression. The higher the better in blue straight lines, and the lower the better in other metrics.}
\end{figure}

%\ref{fig:singlePolicy}
%~\ref{fig:multiPolicy}
We use four different metrics to measure system performance: avg. reward, learning time, avg. task success, and avg. dialogue length. The higher the better in the first and third, and the lower the better in the second and fourth. Figure 2 shows learning curves for the baseline DQN-based system, and Figure 3 shows learning curves for the proposed NDQN-based system. Both the baseline and proposed system report results over 150K learning steps (about 8700 dialogues). Our results report that training multi-domain systems using a single policy is twofold harder than using a multi-policy approach. First, this is evidenced by the fact that the baseline policies do not improve over time, and the policies with the proposed method do. This is presumably due to the abstraction exhibited in the multi-policy approach---more focused system actions rather than interleaving them across domains. Second, our proposed system also learned 4.6 times faster than the baseline, which was accelerated further to 4.7 times faster by using compressed inputs\footnote{Ran on Intel Core i5-3210M CPU $@$ 2.50GHz x 4; 8GiB DDR4 RAM $@$ 2400MHz.}. By applying synonymization we are able to use a smaller vocabulary when training and then a much wider vocabulary at runtime, which adds robustness in the presence of unseen dialogues. These results show indication of better scalability for NDQN to multiple domains. 
%\footnote{The task success curve in the baseline is under investigation---it should improve over time.}
%It can be noted that the compressed inputs represent 
%\footnote{We retrained the baseline policy with a larger number of nodes in the first hidden layer and obtained the same results but with longer training times.}

%In addition, initial results of the SVM-based domain transition classifier shows a 87.5\% classification accuracy on user-simulated data. We aim to improve upon this result in future experiments as more realistic user data is obtained. %Further experiments comparing action-selection strategies and learning rate strategies can be found in Appendixes~\ref{AppendixB} and Appendix~\ref{AppendixC}. 
Although the currently generated dialogues using the proposed method seem reasonable, a natural question to ask is {\it How good (qualitatively speaking) are the trained policies?} This question will be answered in an evaluation reported in future work. 
%Further experiments comparing $e$-greedy against Thomson sampling can be found in Appendix~\ref{AppendixB}, and a comparison of  learning rate strategies can be found in Appendix~\ref{AppendixC}. 

\begin{figure}%[ht!]
     \begin{center}
        \subfigure[Meta Domain: (left) without input compression, (right) with input compression]{%
            %\label{fig:first}
            \includegraphics[width=0.51\textwidth]{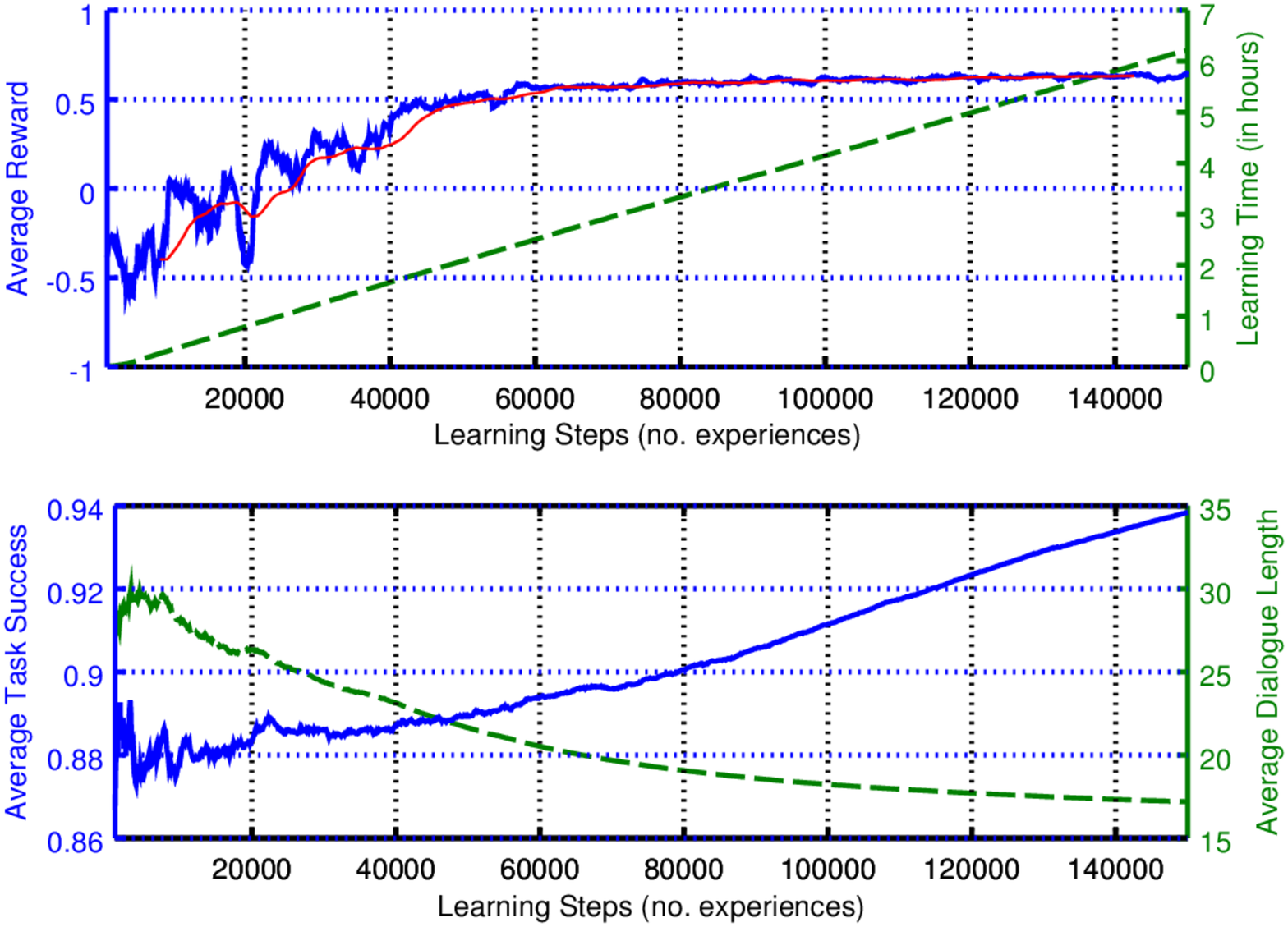}
		\hspace{-0.25cm}
            \includegraphics[width=0.51\textwidth]{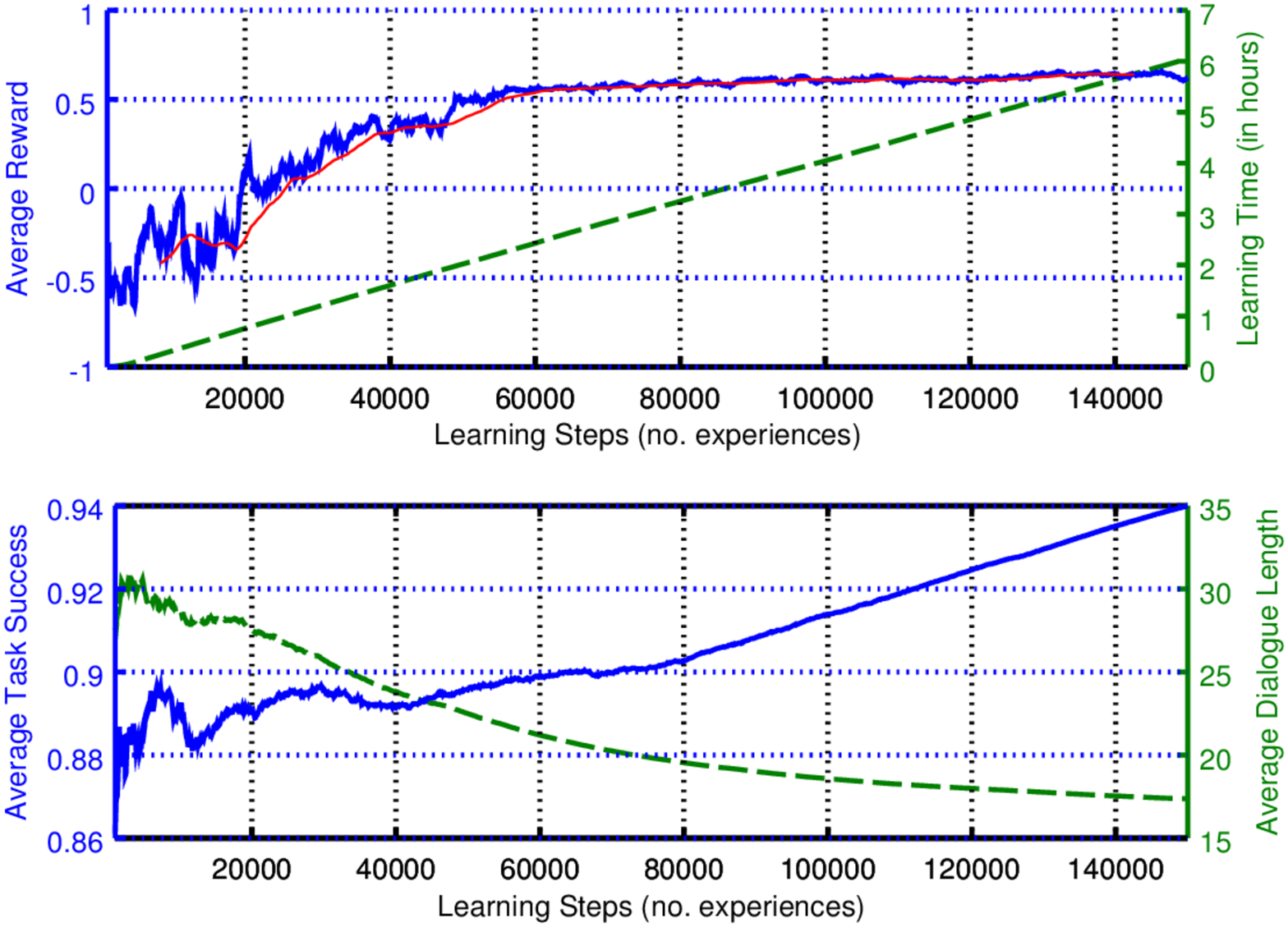}
        }\\
        \vskip30pt
        \subfigure[Restaurants Domain: (left) without input compression, (right) with input compression]{%
           %\label{fig:second}
           \includegraphics[width=0.51\textwidth]{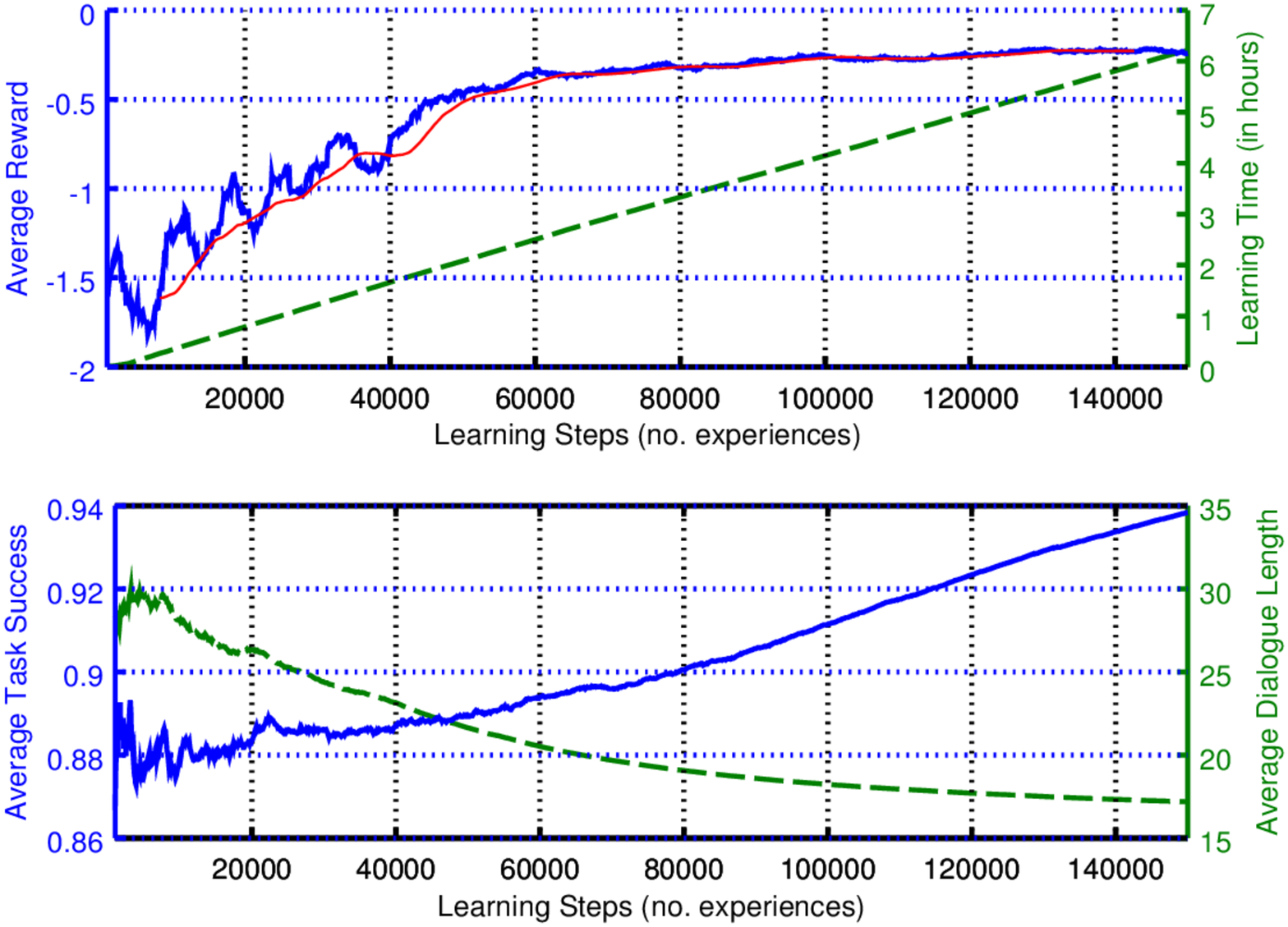}
           \includegraphics[width=0.51\textwidth]{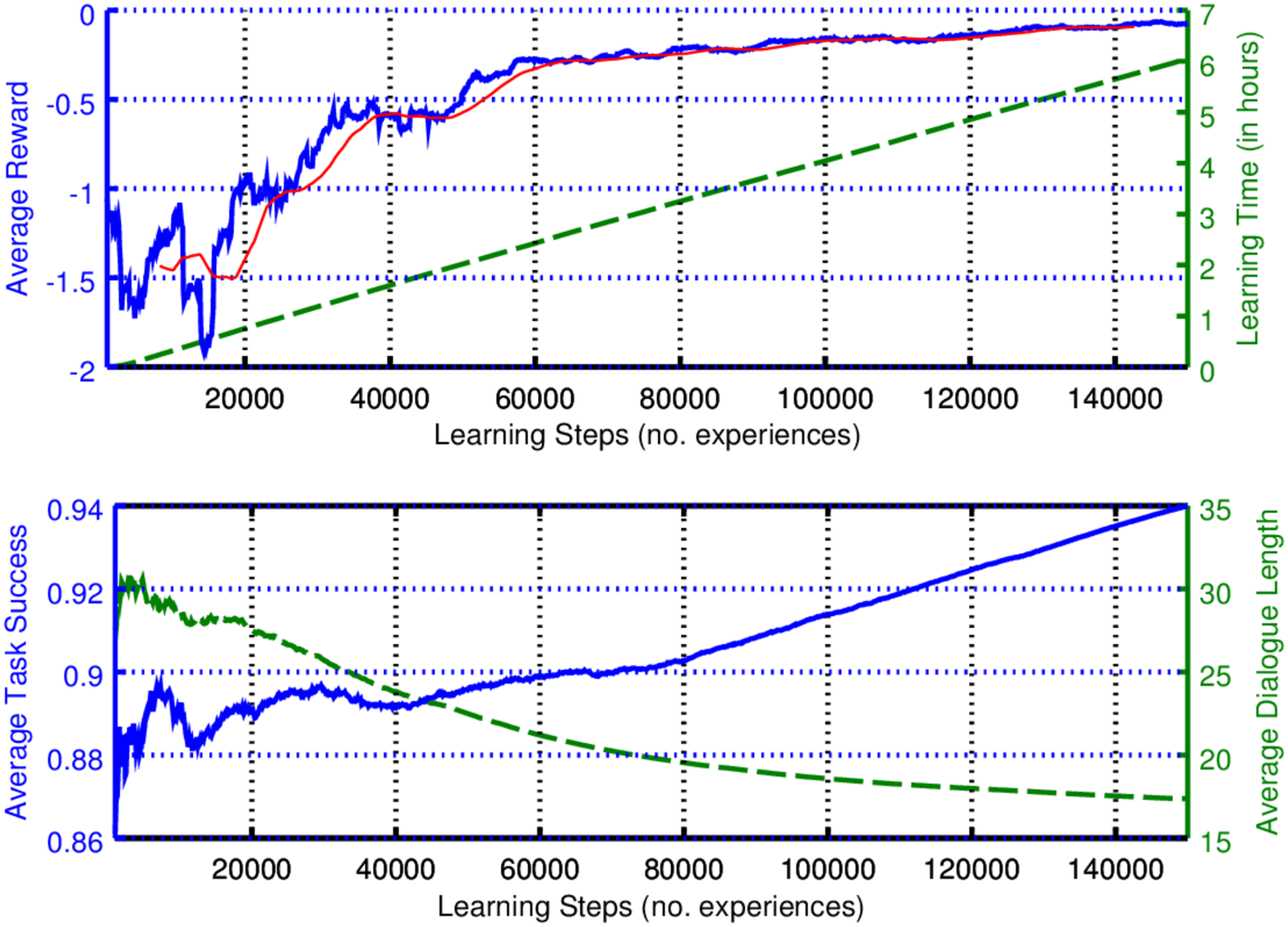}
        }\\
        \vskip30pt
        \subfigure[Hotels Domain: (left) without input compression, (right) with input compression]{%
            %\label{fig:third}
            \includegraphics[width=0.51\textwidth]{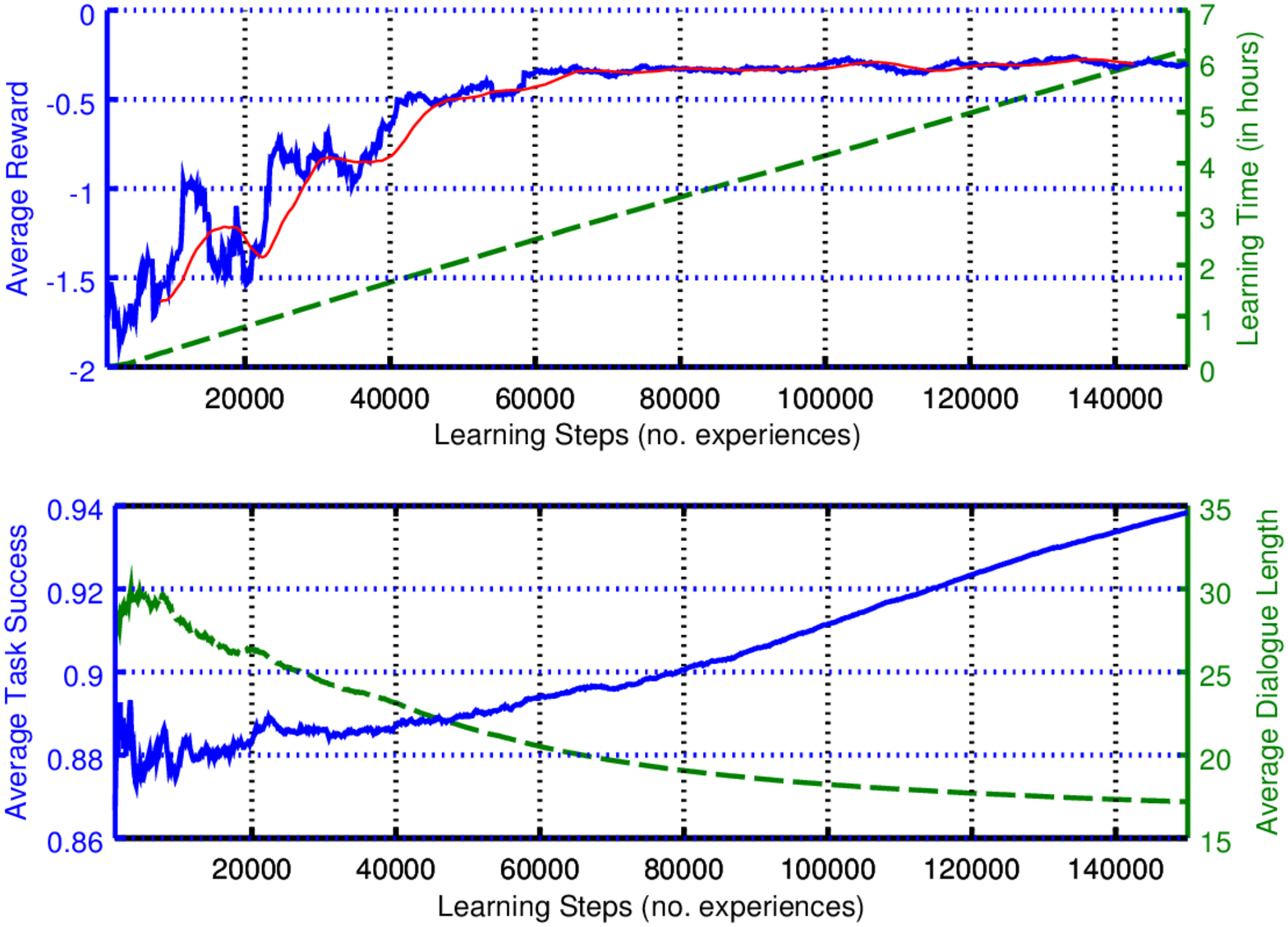}
            \includegraphics[width=0.51\textwidth]{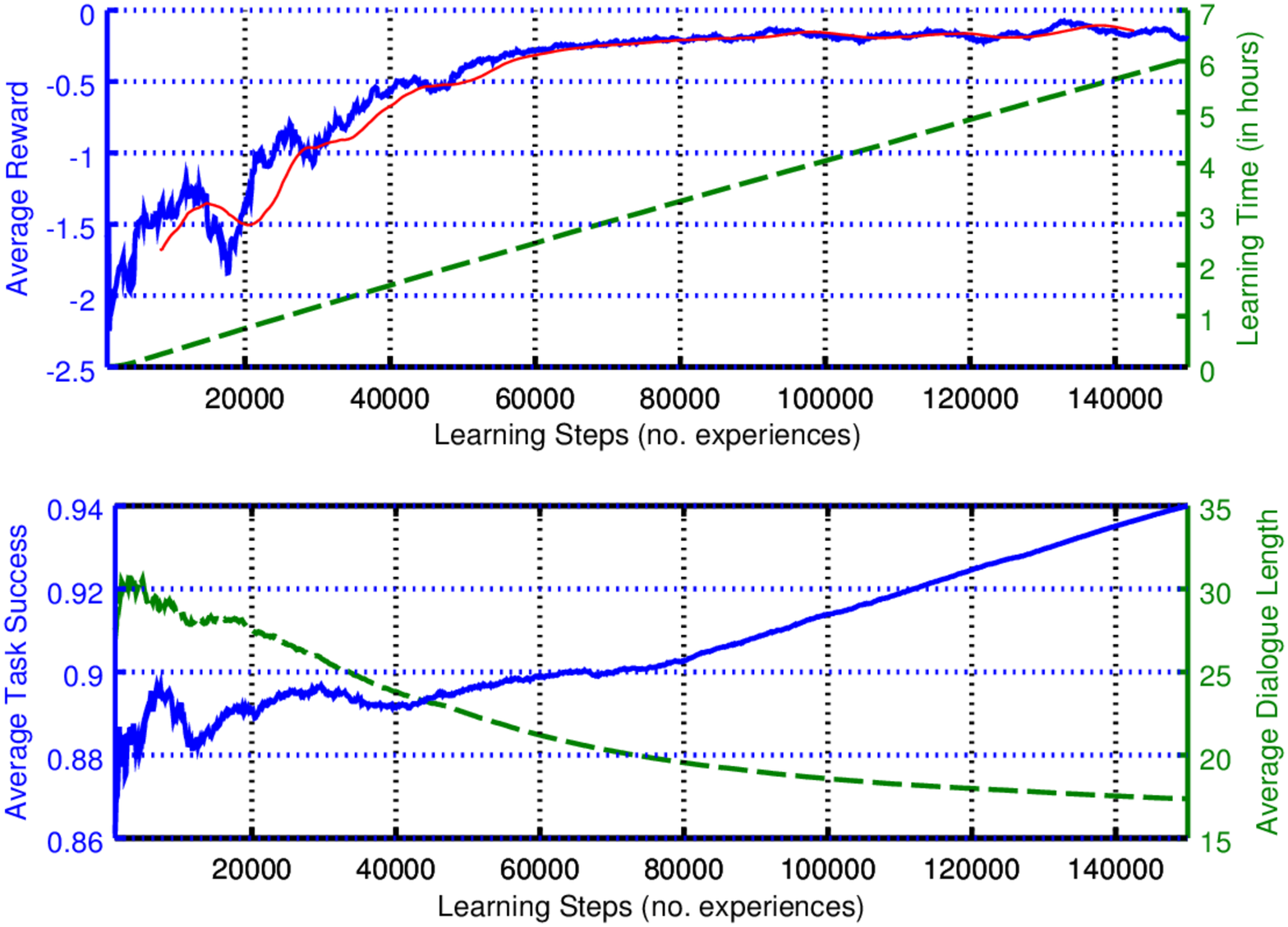}
        }
    \end{center}
        \vskip20pt
   \label{fig:multiPolicy}
   \caption{Learning curves of the proposed system using simultaneous policy learning with the proposed method. The higher the better in  avg. reward and avg. task success, and the lower the better in other metrics. The plots on the left correspond to our proposed system with word-based features, and the plots on the right correspond to our proposed system with delexicalised inputs as features. The latter plots show no performance degradation despite of using more compact state representations.}
\end{figure}

\begin{table}[h!]
\centering
\begin{tabular}[c]{ |p{5cm}||c||c| }
%\begin{center}

 \hline
 %\multicolumn{3}{|c|}{Country List} \\
 %\hline
 Method / System & Baseline (DQN) & Proposed (NDQN) \\
 \hline
Without input compression   &  28.57 hrs    & 6.21 hrs   \\
With input compression   & 16.63 hrs    & 6.05 hrs   \\
 \hline
%\end{center}
\end{tabular}
\caption{Learning times of the baseline and proposed systems} \label{tab:learningTimes}
\end{table}

\section{Conclusion and Future Work}
We have described a method for training multi-domain dialogue systems in a more scalable way than traditional deep reinforcement learning, e.g. using the DQN method. The proposed method uses a Network of DQN (NDQN) agents in order to train specialised agents, and compression of input features. Experimental results using simulations report that the proposed method (NDQN) can train policies faster and more effectively than DQN---almost 5 times faster than DQNs. In addition, our results report that input compression contribute faster learning without performance degradation across metrics. These dialogue agents are the first applications of deep reinforcement learning to optimising the behaviour of multi-domain human-machine dialogues.

%Future work includes a data collection of speech-based dialogues using the {\it MultiDS} App with recruited participants, retraining/improving the multi-domain dialogue policies, and evaluating the retrained policies with crowdsourced users.

In future work we plan to carry out a data collection of speech-based interactions with recruited participants in order to evaluate and/or retrain our multi-domain dialogue policies with a larger dataset. Other future work includes applying and refining the proposed framework to larger sets of domains, and applying it to multimodal interaction in different types of devices and machines.

\appendix
\section{Example Demonstration Dialogue} 
\label{App:AppendixA}
We aim for a data-efficient system, trainable from a small set of demonstrations. This dialogue is one of the 12 manually-specified seed demonstrations for inducing the system's features (words) and actions (dialogue acts), and for training classifiers to predict the most likely actions given the previous system and user responses. The latter were used by the DRL agents described in Section~\ref{sec:method} for inducing their dialogue behaviour according to the learning environment described in Section~\ref{sec:multids}. Square brackets denote system actions, double quotes denote system or user responses, and words beginning with an underscore encapsulate live web-based information according to the user requests. %During testing, our system retrieveslive information from \url{http://www.bookatable.co.uk} and \url{www.reservetravel.com}.

\begin{table*}[h!]
\small
\centering
\begin{tabular}{|c|l|l|l|}
\hline
\bf Domain & \bf Agent & \bf [Action]"Verbalisation" \\
\hline 
%\hline 
Meta & SYS & [Salutation(greeting)]"Welcome to MultiDS!"\\
 &     & [Provide(intro)]"You may ask for information about hotels and restaurants."\\
 &     & [Request(hmihy)]"How can I help you?"\\
 & USR & {\it "I am looking for a hotel in edinburgh on the 2nd of january for 2 nights"}\\
\hline
Hotels & SYS & [Apology(h\_nights)]"Sorry, how long?"\\
 & USR & {\it "2 nights"}\\
 & SYS & [ExpConfirm(h\_day=\$h\_day,h\_month=\$h\_month,h\_nights=\$h\_nights)]\\
 &     & "Did you say on the 2nd of january for 2 nights?"\\
 & USR & {\it"yes i did"}\\
 & SYS & [ExpConfirm(h\_city=\$h\_city)]"Did you say in edinburgh?"\\
 & USR & {\it"i did"}\\
 & SYS & [Retrieve(h\_info)]"Let me see."\\
 &     & [Provide(h\_info)]"I found the following hotels. \_hotels."\\
\hline
Meta &     & [AskFor(h\_more)]"Anything else?"\\
 & USR & {\it "I am looking for cheap japanese food in the centre of town"}\\
\hline
Restaurants & SYS & [Apology(r\_area)]"Sorry, what area?"\\
 & USR & {\it"In the centre"}\\
 & SYS & [ExpConfirm(r\_price=\$r\_price,r\_area=\$r\_area)]\\
 &     & "Did you say cheap food in the centre?"\\
 & USR & {\it"yes i did"}\\
 & SYS & [ExpConfirm(r\_food=\$r\_food)]"Did you say japanese food?"\\
 & USR & {\it"i did"}\\
 & SYS & [Retrieve(h\_info)]"Let me see."\\
 &     & [Provide(h\_info)]"I found the following restaurants. \_res."\\
\hline
Meta &     & Anything else?\\
 & USR & {\it "no"}\\
 & SYS & [Salutation(closing)]"Okay, talk to you soon. Bye!"\\
\hline
\end{tabular}
%\caption{Example demonstration dialogue. Square brackets denote system actions, double quotes denote system or user responses, and words beginning with an underscore encapsulate live web-based information according to the user requests. During testing, our system retrieveslive information from \url{http://www.bookatable.co.uk} and \url{www.reservetravel.com}.}\label{exampleDialogue}
\end{table*}

\small
\bibliographystyle{abbrv}
\bibliography{multids}

\end{document}